# A Review of Challenges and Opportunities in Machine Learning for Health

Marzyeh Ghassemi, PhD[1], Tristan Naumann, PhD[2], Peter Schulam, PhD[3], Andrew L. Beam, PhD[4], Irene Y. Chen, SM[5], Rajesh Ranganath, PhD[6]
[1]University of Toronto and Vector Institute, Toronto, Canada; [2]Microsoft Research, Redmond, WA, USA; [3]Johns Hopkins University, Baltimore, MD, USA; [4]Harvard School of Public Health, Boston, MA, USA; [5]Massachusetts Institute of Technology, Cambridge, MA, USA; [6]New York University, New York, NY, USA;

**Abstract**

*Modern electronic health records (EHRs) provide data to answer clinically meaningful questions. The growing data in EHRs makes healthcare ripe for the use of machine learning. However, learning in a clinical setting presents unique challenges that complicate the use of common machine learning methodologies. For example, diseases in EHRs are poorly labeled, conditions can encompass multiple underlying endotypes, and healthy individuals are underrepresented. This article serves as a primer to illuminate these challenges and highlights opportunities for members of the machine learning community to contribute to healthcare.*

## Introduction

Health problems impact human lives. During medical care, health providers collect clinical data about each particular patient, and leverage knowledge from the general population, to determine how to treat that patient. Data thus plays a fundamental role in addressing health problems, and improved information is crucial to improving patient care.

Using data, machine learning has driven advances in many domains including computer vision, natural language processing (NLP), and automatic speech recognition (ASR) to deliver powerful systems (e.g., driverless cars, voice activated personal assistants, automated translation). Machine learning's ability to extract information from data, paired with the centrality of data in healthcare, makes research in machine learning for healthcare crucial.

Interest in machine learning for healthcare has grown immensely, including work in diagnosing diabetic retinopathy[1], detecting lymph node metastases from breast pathology[2], autism subtyping by clustering comorbidities[3], and large-scale phenotyping from observational data[4]. Despite these advances, the direct application of machine learning to healthcare remains fraught with pitfalls. Many of these challenges stem from the nominal goal in healthcare to make personalized predictions using data generated and managed via the medical system, where data collection's primary purpose is to support care, rather than facilitate subsequent analysis.

Existing reviews of machine learning in the medical space have focused narrowly on biomedical applications[5], deep learning tasks well suited for healthcare[6], the need for transparency[7], and use of big data in precision medicine[8]. Here, we emphasize the broad opportunities present in machine learning for healthcare and the careful considerations that must be made. We focus on the electronic health record (EHR), which documents the process of healthcare delivery and operational needs such as tracking care and revenue cycle management (i.e., billing and payments). While we choose to focus on the inpatient setting as the majority of machine learning projects currently focus on this data-rich environment, we note that clinical data is heterogeneous, and comes in a variety of forms that can be relevant to understanding patient health[9].

In this work, we cover the unique technical challenges that should be considered in machine learning systems for healthcare tasks, especially as performance between trained models and human experts narrows[10]. Failure to carefully consider these challenges can hinder the validity and utility of machine learning for healthcare. We present a hierarchy of clinical opportunities, organized into the following general categories: automating clinical tasks, providing clinical support, and expanding clinical capacities. We conclude by outlining the opportunities for research in machine learning that have particular relevance in healthcare: accommodating shifts in data sources and mechanisms, ensuring models are interpretable, and identifying good representations.

## Unique Technical Challenges in Healthcare Tasks

In tackling healthcare tasks, there are factors that should be considered carefully in the design and evaluation of machine learning projects: causality, missingness, and outcome definition. These considerations are important across both modeling frameworks (e.g., supervised vs. unsupervised), and learning targets (e.g., classification vs. regression).

*Understanding Causality is Key*

Many of the most important and exciting problems in healthcare require algorithms that can answer *causal*, "what if?" questions about what will happen if a doctor administers a treatment[11]. These questions are beyond the reach of classical machine learning algorithms because they require a formal model of interventions. To address this class of problems, we need to reason about and learn from data through the lens of *causal models* (e.g., see prior work[12,13]). Learning from data to answer causal questions is most challenging when the data are collected *observationally*; that is, it *may* have been influenced by the actions of an agent whose policy for choosing actions is not known.

In healthcare, learning is done almost exclusively using observational data, which poses a number of challenges to building models that can answer causal questions. For instance, Simpson's paradox describes the observation that the relationship between two variables can change directions if more information is included in the model[12]. To better understand this issue, consider prior work in which researchers found that asthmatic patients who were admitted to the hospital for pneumonia were more aggressively treated for the infection, lowering the subpopulation mortality rate[14]. A model that predicts death from asthma will learn that asthma is protective. If, however, an additional variable to account for the level of care is included, the model may instead find that having asthma increases the risk of death. This example demonstrates that causal models are not only useful to evaluate treatments, but can also help to build reliable predictive models that do not make harmful predictions using relationships caused by treatment policies in the training data[11]. To account for these challenges, strong assumptions must be made that cannot be statistically checked or validated; i.e., gathering more data will not help[15]. The shortcomings of classical statistical learning when answering causal questions are discussed in greater detail in prior work[16].

*Models in Health Must Consider Missingness*

Even if all important variables are included in a healthcare dataset, it is likely that many observations will be missing. Truly complete data is often impractical due to cost and volume. Learning from incomplete, or *missing*, data has received little attention in the machine learning community (although exceptions exist[17-19]), but is an actively studied topic in statistics[20]. Because healthcare is a dynamic process where vitals are measured and labs are ordered over time by doctors in response to previous observations there are strong dependencies between what variables are measured and their values, which must be carefully accounted for to avoid biased results (e.g. see prior work[21] Appendix B).

There are three widely accepted classifications of *missing data mechanisms*; i.e., the measurement mechanism determining whether a value is recorded or not[22]. The first, *missing completely at random* (MCAR), posits a fixed probability of missingness. In this case, dropping incomplete observations---known as complete case analysis---is commonly used (albeit naively), and will lead to unbiased results. Second, the data may be *missing at random* (MAR), where the probability of missingness is random conditional on the observed variables. In this case, common methods include re-weighting data with methods like inverse probability of censoring weighting or using multiple imputations to in-fill[23,24]. Finally, data may be *missing not at random* (MNAR), where the probability of missingness depends on the missing variable itself, or other missing and unobserved variables.

*Sources of missingness must be carefully understood.* Sources of missingness should be carefully examined before deploying a learning algorithm. For example, lab measurements are typically ordered as part of a diagnostic work-up, meaning that the presence of a datapoint conveys information about the patient's state. Consider a hospital where clinical staff measures patient lactate level. If a power outage led to a set of lactate levels being lost, the data are MCAR. If nurses are less likely to measure lactate levels in patients with traumatic injury, and we record whether patients were admitted with trauma, the data are MAR. However, if nurses are less likely to measure lactate levels when believed to be already, then the lactate measures themselves are MNAR, and the measurement of the signal itself is meaningful. The key feature of missing data is that there may be information conveyed by the absence of an observation, and ignoring this dependence may lead to models that make incorrect, and even harmful, predictions.

*Include missingness in the model.* Including missingness indicators provides the most information for making predictions[25]. However, learning models without an appropriate model of missingness leads to issues such as inaccurate assessment of feature importance and models that are brittle to changes in measurement practices. For example, troponin-T is commonly measured only when a myocardial infarction is considered probable. A model learned by treating troponin-T as MCAR would likely overpredict the rate of myocardial infarction on data where troponin-T was more regularly measured. A model trained with MAR troponin values would be more robust to this.

*Missingness can reflect human biases.* We note that data may also be missing because of differences in access, practice, or recording that reflects societal biases[26-27]. Models trained on such data may in turn exhibit unfair performance for some populations if a machine learning practitioner is unaware of this underlying variation; thus, checking models across groups is important[28].

*Make Careful Choices in Defining Outcomes*

Obtaining reliable outcomes for learning is an important step in defining tasks. Outcomes are often used to create the gold-standard labels needed for supervised prediction tasks, but are crucial in other settings as well, e.g., to ensure well-defined cohorts in a clustering task. There are three key factors to consider with outcome definitions: creating reliable outcomes, understanding the relevance of an outcome clinically, and the subtlety of label leakage.

*Create reliable outcomes from heterogeneous source data.* Multiple data sources should be considered when creating labels because EHRs often lack precise structured labels. Or, in some cases, structured labels may unreliable[29]. For example, a diagnostic clinical code for pneumonia could mean a patient was *screened* for pneumonia rather than that they *actually had* pneumonia. Machine learning methods to pool different data types and obtain a more reliable label is known as *phenotyping*[30], and is an important subfield of machine learning in healthcare[31,32]. Recent work has emphasized the need to integrate the rich information available in clinical notes, and building natural language processing pipelines to extract information from unstructured clinical text accurately is an active subject of research[33].

*Understand the outcome in context of a healthcare system.* Medical definitions are working models using current scientific understanding of a disease. As understanding evolves, so does the definition. The implication for machine learning is that good predictive performance on labels based on such definitions is only as good as the underlying criteria. For example, acute kidney injury (AKI) is an important critical illness with two recent definitions: RIFLE[34] and KDIGO[35]. Similarly, it is tempting to use the actual behavior of clinicians as labels, but it is important to remember that they may not be correct. For example, work that targets prediction of clinical actions must carefully consider whether the treatments are good labels, and whether ``incorrect predictions'' are in fact forecasting treatments that would have been given by other clinicians rather than treatments that would optimally treat the patient[36-38].

*Beware of label leakage.* The information collected in an individual's hospital encounter is tightly coupled across time, and this can result in information about the targeted task outcome leaking back into features. While exploiting such relationships between features and targets is a goal of learning, information leakage can render a prediction meaningless. Consider predicting mortality of hospital patients using all available data up until their time of death. Such a task could lead to a pathological prediction rule—"if the ventilator is turned off in the preceding hour, predict death." This commonly happens when patients and their families decide to withdraw care at a terminal stage of illness. A machine learning algorithm trained naively on this signal would have high predictive performance by nearly any metric, yet absolutely no clinical utility. High capacity neural models examining the entire EHR may also be subject to label leakage[39]; for example, a final diagnostic label could leverage labels that appear within clinicians' initial hypotheses, confirming clinical consistency in hypotheses, rather than predicting a diagnosis based on relevant data.

**Addressing a Hierarchy of Healthcare Opportunities**

There are many high-impact opportunities in healthcare. Before fitting models, goals should be clearly identified and validated as worth solving. Here, we frame potential healthcare opportunities into three high-level categories: automating clinical tasks, providing clinical support, and expanding clinical capacities. The details of how a technical solution is deployed can change its intent, and it is therefore crucial to engage clinical stakeholders early on.

*Clinical Task Automation: Automating clinical tasks during diagnosis and treatment*

There are many tasks currently performed by clinicians that present low-hanging fruit for machine learning researchers. Clinical task automation encompasses a class of work that clinicians currently do. These tasks are well-defined (i.e., known input and output spaces), and thus require the least amount of domain adaptation and investment. The evaluation of task replacement is also straightforward—performance can be measured against existing standards. We emphasize that algorithms should not replace clinical staff, but rather be used to optimize the clinical workflow. Clinical roles will likely evolve as these techniques improve, empowering staff to spend more time with patients[40].

*Automating medical image evaluation.* Medical imaging is a natural opportunity for machine learning because clinicians undergo intensive training to map from a fixed input space (e.g. the images) to an output (e.g. the diagnosis). There have been several recent successes in applying deep learning to medical imaging tasks. For example, physician-level parity in the detection of diabetic retinopathy[1], distinguishing between malignant and non-malignant skin lesions using in dermatoscopic images[41], detecting lymph node metastases from breast pathology slides[2], and detecting hip fractures in x-ray images[42].

*Automating routine processes.* Similarly, automating routine clinical processes stands to reduce the burden placed on clinical staff[43]. For example, prioritizing triage order in the emergency department is often left to staff[44], but could be done algorithmically. Likewise, summarizing the contents of patients' medical records[45] is a time-consuming, but valuable, task. For example, when hospital staff are unsure about patient's disease status, they may call for an infectious disease consultation in which a specialist meticulously reviews all the available patient data, and manually summarizes disparate sources into a series of recommended tests and treatments[44]. Researchers can leverage work from natural language generation working towards generating summaries from structured her data, saving much needed time[47].

### Clinical Support and Augmentation: Optimizing clinical decision and practice support

Another set of opportunities focus on supporting and augmenting care. Rather than replacing a well-defined task, support requires understanding clinical pain points and working with clinical staff to understand appropriate input data, output targets and evaluation functions. Opportunities for support focus on work that often suffers due to real-world constraints on time and resources, often leading to information loss and errors (e.g., mistaken patient identification, flawed interpretations, or incorrect recall[48]). In this setting, it is appropriate to evaluate how models improve downstream outcomes in tandem with clinical input rather than head-to-head comparisons to clinical staff[49].

*Standardizing clinical processes.* Variations in clinical training and experience lead to ranges of treatment choices that may not be optimal for targeting the underlying maladies of a patient's state. For example, clinical staff may be unsure which medication sets or doses are most appropriate for a patient[50]. To support such needs, past work has examined recommending both standardized order sets to help care staff quickly assess what medications they may have missed[51] and default dosages to avoid dangerous dosing[52]. We note that automating a static, existing, protocol is significantly easier than full decision support for complex clinical guidelines that may change over time, and there is potentially much value in both in-hospital and ambulatory decision support[53], and at-home support[54].

*Integrating fragmented records.* Finite resources can also lead to a lack of communication and coordination, affecting patient care. For example, it can take years to identify domestic abuse survivors because any single clinical visit in isolation may be consistent with other causes (e.g. an admission for bruising is consistent with a spontaneous fall). Only a thorough review of a patient's record will demonstrate the pattern of repeated admissions and other indicators (e.g., partners' alcohol abuse[55]). While possible without support systems in principle, machine learning can be a powerful tool, e.g., identifying domestic abuse up to 30 months in advance of the healthcare system[56].

### Expanding Clinical Capacities: New horizons in screening, diagnosis and treatment.

As healthcare records become increasingly digitized, clinicians are faced with an ever-increasing amount of novel data for patients and populations. The result is an opportunity to give the healthcare system a new set of capacities to deliver healthcare in better and smarter ways. Importantly, creating new capacities requires the most involvement with clinical collaborators, and impact should be measured both in innovation and clinical value.

*Expanding the coverage of evidence.* While healthcare is an inherently data-driven field, most clinicians operate with limited evidence guiding their decisions. Randomized trials estimate average treatment effects for a trial population, but numerous day-to-day decisions are not based on high-quality randomized control trials (RCTs)[57]. For example, the majority of commonly used ICU treatments are not rigorously empirically validated[58]; some analysts have estimated that only 10-20%[59] of treatments are backed by an RCT. Even in settings where an RCT exists, the trial population tends to be a narrow, focused subgroup defined by the trial's inclusion criteria[60], but that cohort may not be representative of the heterogeneous population to which trial results are then applied[61]. Finally, RCT results cannot reflect the complexity of treatment variation because, in practice, patient care plans are highly individualized. Prior work found that approximately 10% of diabetes and depression patients and almost 25% of hypertension patients had a unique treatment pathway (i.e., zero nearest neighbors) in a cohort of 250 million patients[62]. One way forward would be to leverage this naturally occurring heterogeneity to design natural experiments[63] that approximate the results of an RCT using fewer resources, thereby allowing a much larger set of clinical questions to be investigated.

*Moving towards continuous behavioral monitoring.* Phenotyping is an important goal in healthcare[30], and *wearable data* provides an ongoing way for devices to collect continuous non-invasive data and provide meaningful classifications or alerts when a patient needs attention. Recent work has focused on predicting such varied outcomes as heart attack from routinely collected data[64]. There are further settings where non-invasive monitoring may be the only practical way to provide detection. For example, automatic fall detection for geriatric patients, or enforcing hand washing compliance in a clinical setting[65]. Importantly, prior challenges identified in label leakage, soft labels, confounding and missingness must be considered carefully. In phenotyping chronic conditions, patients are often already being treated, and so early *detection* may sometimes amount to identifying an existing treatment.

*Precision medicine for early individualized treatment.* Precision medicine seeks to individualize the treatment of each patient; this is particularly important for syndromes—conditions defined by a collection of symptoms whose causes are unknown[66]. For instance, acute kidney injury (AKI) is defined by a collection of symptoms characterizing kidney failure, not an underlying cause[67]. Two individuals may have developed AKI for different reasons because there are many reasons that kidneys can fail. More measurements of the two individuals could reveal the difference in cause, which may in turn suggest alternative treatment strategies[68]. By personalizing over time, one can learn individual-specific treatment effects that address the cause of the syndrome[11,69,70] in a particular individual. This relates to the ideas from "*N=1*" crossover studies in experimental design[72]. Personalized treatment is enabled by growing repositories of *longitudinal* data, where long-term progressions of an individual's health are available[73-76].

**Opportunities for New Research in Machine Learning**

Addressing the hierarchy of opportunities in healthcare creates numerous opportunities for innovation. Importantly, clinical staff and machine learning researchers often have complementary skills, and many high-impact problems can only be tackled by collaborative efforts. We note several promising directions of research, specifically highlighting those that address issues of data non-stationarity, model interpretability, and discovering appropriate representations.

*Accommodating Data and Practice Non-stationarity in Learning and Deployment*

In most existing work, models are trained on the largest dataset possible and assumed to be fit for deployment, i.e., models do not keep learning. This is problematic in clinical settings, because patient populations and recommended treatment procedures will change over time, resulting in degraded predictive performance as the statistical properties of the target change. For example, clinicians previously assumed that estrogen was cardioprotective in menopausal women[77] and hormone therapy was routinely prescribed as a preventative measure until large trials reported either no benefit or an increase in adverse cardiac events[78]. In developing new models for healthcare, models must be made robust to these changes, or acknowledge their mis-calibration for the new population[79].

*Internal Validity - Shift over time.* In a notable example of concept drift, Google Flu Trends persistently overestimated flu due to shifts in search behaviors[80]. The initial model was a great success, leveraging Google search data to predict flu incidence; however, without update the model began to overestimate flu incidence in subsequent years as user search behaviors had shifted. While the drift was unintentional, the example motivates the need for models that continuously update.

*External Validity - Shift over sources.* There is also no reason to believe *a priori* that a model learned from one hospital will generalize to a new one. Many factors impact generalizability, including local hospital practices, different patient populations, available equipment, and even the specific kind of EHR each uses—transitions from one EHR to another create non-obvious feature mapping problems[81,82]. This issue will remain until infrastructure to easily test across multiple sites becomes prevalent. The absence of such standardization creates opportunities with respect to data normalization and the development of models that are robust to differences in data collection at different sites[83-84].

*Creating models robust to feedback loops.* Models that learn from existing clinical practice are susceptible to amplifying the biases endemic to modern healthcare. While not yet observed in healthcare, such feedback loops have been noted in the deployment of predictive policing[85]. Such biases reflected in deployed predictions can propagate into future training data, effectively creating a feedback loop that causes further bias[86]. Work in algorithmic fairness should be considered as it motivates the need for systems that are sufficiently aware that they can alert us to such unwanted behavior. Additionally, models trained on EHR will learn from existing procedures, rather than represent formally best-practice protocols, which may be problematic given low manual compliance in much clinical practice[87].

*Creating interpretable models and recommendations*

In a clinical setting, black box methods present new challenges. Traditionally, quantitative training has not been emphasized in a physician's extensive medical training[88] and most physicians do not have a robust understanding of rubrics such as positive predictive value[89]. Models cannot be deployed "in the wild" at a low cost, and clinical staff must justify deviations in treatment to satisfy both clinical and legal requirements.

*Defining what interpretibility means.* There are many possible ways to interpretability, e.g., through feature space minimization, model regularization, or a preference for particular classes of models that have well-known post-hoc analysis methods[90]. For example, providing a posterior distribution over possible decision lists[91]. Such lists can provide a natural way for clinicians to think about the relative risks of their patient's condition.

*Moving from interpretation to justification.* In other domains, many forms of interpretability rely on human expertise, e.g., a model may highlight a single sentence from a user review ("The coffee is wonderful.") as the rationale for a review prediction. Clinicians are unlikely to have a similar contextual framework, and it is unlikely to be obvious what a particular pattern of lab measurements that maximally activates a model means, biologically or clinically[45]. We argue that models should instead provide ``justifiability''; beyond explaining a specific prediction, models should strive towards justifying the predictive path itself. For example, recent work has proposed that locally-interpretable results can be presented for each individual prediction[92]. Another possibility is learning influence functions to trace a model's prediction through the learning algorithm and back to its training data, thereby identifying training points most responsible for a given prediction[93]. Outside of reassuring end users, justifying a machine learning algorithm's decision is also important for security concerns, as medicine may be uniquely vulnerable to "adversarial attacks"[94,95].

*Adding interaction to machine learning and evaluation.* Machine learning work in healthcare has an opportunity to create systems that interact and collaborate with human experts. Clinical staff provide more than their expertise; they also act as caregivers, and empathy is recognized as an important element of clinical practice[96,97]. Building *collaborative* systems can leverage the complementary strengths of physicians and learning systems. There are many threads of research in the machine learning literature that can serve as a foundation for such systems. In active learning, for example, the goal is to leverage an *oracle* in order to learn using fewer samples[98]. Apprenticeship learning is another related set of ideas[99]. The study of collaborative systems, where the human and machine work together, is still in its early stages. Examples of such systems include content creation algorithms that alternate back and forth between human and machine proposals[108], and intelligent, data-driven operations for drawing software[101].

### Identifying Representations in a Large, Multi-source, Network

Representation learning has prompted great advances in machine learning; for example, the lower dimensional, qualitatively meaningful representations of imaging datasets learned by convolutional neural networks. Healthcare data lacks such obviously natural structures, and investigations into appropriate representations should include multi-source integration, and learning domain appropriate representations.

*Integrating predictions from multi-source high-dimensional data.* Individual patient data has ballooned to include many sources and modalities, making integration more challenging in systems currently struggling with overload[88,102]. Using high-dimensional data to make well-calibrated predictions of established risks is a way that researchers can contribute. For example, inferring drug-resistance status in tuberculosis from whole-genome sequencing data[103], predicting cardiovascular risk from EHR data[64,104], readmission and mortality risk from longitudinal EHR data[39,105], and predicting cardiovascular risk from retinal images[106]. New memory models are needed for sequence prediction tasks, because such records are not evenly spaced, can cover very long durations, and early events can affect patient state many years later; early work in this space has targeted learning individual variable decays in EHR data with missingness[107].

*Learning meaningful representations for the domain.* Learning meaningful state representations that provide good predictive performance on diverse tasks, and account for conditional relationships of interest, is an important area. Dealing with representations explicitly may be advantageous because they can conveniently express general priors that are not specific to a single predictive task[108], and this is particularly important for zero-shot learning[109] in unseen categories. There are several potential opportunities to address for representations for a clinical setting. First, a single patient input (e.g., physiological data) can correspond to many possible correct outputs (e.g., diagnostic codes), and this must be possible in the representations we explore. Additionally, there is likely value in incorporating structure and domain knowledge into representation learning. There has also been some initial exploration into learning representations of other data types that simultaneously capture hierarchy and similarity[110-112].

### Conclusion

In this work, we give a practical guide that researchers can engage with as they begin work in healthcare. We emphasize that there are many opportunities for machine learning researchers to collaborate with clinical staff, and encourage researchers to engage with clinical experts early on as they identify and tackle important problems. Such efforts may lead to models that are clinically useful and operationally feasible[113].

### Acknowledgements

ALB is supported by NIH award 7K01HL141771-02. MG is funded in part by Microsoft Research, a CIFAR AI Chair at the Vector Institute, a Canada Research Council Chair, and an NSERC Discovery Grant.

# References

1. Gulshan V, Peng L, Coram M, Stumpe MC, Wu D, Narayanaswamy A, et al. Development and validation of a deep learning algorithm for detection of diabetic retinopathy in retinal fundus photographs. Jama. 2016;316(22):2402–2410.
2. Golden JA. Deep Learning Algorithms for Detection of Lymph Node Metastases From Breast Cancer: Helping Artificial Intelligence Be Seen. Jama. 2017;318(22):2184–2186.
3. Doshi-Velez F, Ge Y, Kohane I. Comorbidity clusters in autism spectrum disorders: an electronic health record time-series analysis. Pediatrics. 2014;133(1):e54–e63.
4. Pivovarov R, Perotte AJ, Grave E, Angiolillo J, Wiggins CH, Elhadad N. Learning probabilistic phenotypes from heterogeneous EHR data. Journal of biomedical informatics. 2015;58:156–165.
5. Baldi P. Deep learning in biomedical data science. Annual Review of Biomedical Data Science. 2018;1:181–205.
6. Esteva A, Robicquet A, Ramsundar B, Kuleshov V, DePristo M, Chou K, et al. A guide to deep learning in healthcare. Nature medicine. 2019;25(1):24.
7. Wainberg M, Merico D, Delong A, Frey BJ. Deep learning in biomedicine. Nature biotechnology. 2018;36(9):829.
8. Beckmann, J. S., & Lew, D. (2016). Reconciling evidence-based medicine and precision medicine in the era of big data: challenges and opportunities. *Genome medicine*, *8*(1), 134.
9. Weber GM, Mandl KD, Kohane IS. Finding the missing link for big biomedical data. Jama. 2014;311(24):2479–2480.
10. Topol EJ. High-performance medicine: the convergence of human and artificial intelligence. Nature medicine. 2019;25(1):44.
11. Schulam P, Saria S. Reliable decision support using counterfactual models. In: Neural Information Processing Systems (NIPS); 2017. .
12. Pearl J. Causality: models, reasoning, and inference. Cambridge University Press; 2009.
13. Hernan MA, Robins JM. Causal inference. CRC Boca Raton, FL; 2010.
14. Cooper GF, Abraham V, Aliferis CF, Aronis JM, Buchanan BG, Caruana R, et al. Predicting dire outcomes of patients with community acquired pneumonia. Journal of biomedical informatics. 2005;38(5):347–366.
15. Greenland S, Robins JM, Pearl J. Confounding and collapsibility in causal inference. Statistical science. 1999;p. 29–46.
16. Pearl J. Theoretical Impediments to Machine Learning With Seven Sparks from the Causal Revolution. arXiv preprint arXiv:180104016. 2018;.
17. Marlin BM, Zemel RS, Roweis ST, Slaney M. Recommender systems, missing data and statistical model estimation. In: International Joint Conference on Artificial Intelligence (IJCAI). vol. 22; 2011. p. 2686.
18. Mohan K, Pearl J, Tian J. Graphical models for inference with missing data. In: Neural Information Processing Systems (NIPS); 2013.
19. McDermott MBA, Yan T, Naumann T, Hunt N, Suresh H, Szolovits P, et al. Semi-supervised Biomedical Translation with Cycle Wasserstein Regression GANs. In: Association for the Advancement of Artificial Intelligence. New Orleans, LA; 2018. .
20. Ding P, Li F. Causal inference: A missing data perspective. Statistical Science. 2018;33(2):214–237.
21. Schulam P, Saria S. Integrative analysis using coupled latent variable models for individualizing prognoses. The Journal of Machine Learning Research (JMLR). 2016;17(1):8244–8278.
22. Little RJA, Rubin DB. Statistical analysis with missing data. vol. 333. John Wiley & Sons; 2014.
23. Robins JM. Robust estimation in sequentially ignorable missing data and causal inference models. In: Proceedings of the American Statistical Association. vol. 1999; 2000. p. 6–10.
24. Robins JM, Rotnitzky A, Scharfstein DO. Sensitivity analysis for selection bias and unmeasured confounding in missing data and causal inference models. In: Statistical models in epidemiology, the environment, and clinical trials. Springer; 2000. p. 1–94.
25. Miscouridou X, Perotte A, Elhadad N, Ranganath R. Deep survival analysis: Nonparametrics and missingness. In: Machine Learning for Healthcare Conference; 2018. p. 244–256.
26. Rajkomar A, Hardt M, Howell MD, Corrado G, Chin MH. Ensuring fairness in machine learning to advance health equity. Annals of internal medicine. 2018;169(12):866–872.
27. Rose S. Machine learning for prediction in electronic health data. JAMA network open. 2018;1(4):e181404–e181404.
28. Chen IY, Szolovits P, Ghassemi M. Can AI Help Reduce Disparities in General Medical and Mental Health Care? AMA Journal of Ethics. 2019;21(2):167–179.
29. O'malley KJ, Cook KF, Price MD, Wildes KR, Hurdle JF, Ashton CM. Measuring diagnoses: ICD code accuracy. Health Services Research. 2005;40(5p2):1620–1639.

30. Richesson RL, Hammond WE, Nahm M, Wixted D, Simon GE, Robinson JG, et al. Electronic health records based phenotyping in next-generation clinical trials: a perspective from the NIH Health Care Systems Collaboratory. Journal of the American Medical Informatics Association. 2013;20(e2):e226–e231.
31. Halpern Y, Horng S, Choi Y, Sontag D. Electronic medical record phenotyping using the anchor and learn framework. Journal of the American Medical Informatics Association. 2016;23(4):731–740.
32. Yu S, Ma Y, Gronsbell J, Cai T, Ananthakrishnan AN, Gainer VS, et al. Enabling phenotypic big data with PheNorm. Journal of the American Medical Informatics Association. 2017;25(1):54–60.
33. Savova GK, Masanz JJ, Ogren PV, Zheng J, Sohn S, Kipper-Schuler KC, et al. Mayo clinical Text Analysis and Knowledge Extraction System (cTAKES): architecture, component evaluation and applications. Journal of the American Medical Informatics Association. 2010;17(5):507–513.
34. Ricci Z, Cruz D, Ronco C. The RIFLE criteria and mortality in acute kidney injury: a systematic review. Kidney International. 2008;73(5):538–546.
35. Khwaja A. KDIGO clinical practice guidelines for acute kidney injury. Nephron Clinical Practice. 2012;120(4):c179–c184.
36. Ghassemi M, Wu M, Feng M, Celi LA, Szolovits P, Doshi-Velez F. Understanding vasopressor intervention and weaning: Risk prediction in a public heterogeneous clinical time series database. Journal of the American Medical Informatics Association. 2016;p. ocw138.
37. Ghassemi M, Wu M, Hughes M, Doshi-Velez F. Predicting Intervention Onset in the ICU with Switching State Space Models. In: Proceedings of the AMIA Summit on Clinical Research Informatics (CRI). vol. 2017. American Medical Informatics Association; 2017. .
38. Suresh H, Hunt N, Johnson A, Celi LA, Szolovits P, Ghassemi M. Clinical Intervention Prediction and Understanding with Deep Neural Networks. In: Machine Learning for Healthcare Conference; 2017. p. 322–337.
39. Rajkomar A, Oren E, Chen K, Dai AM, Hajaj N, Liu PJ, et al. Scalable and accurate deep learning for electronic health records. arXiv preprint arXiv:180107860. 2018;.
40. Jha S, Topol EJ. Adapting to artificial intelligence: radiologists and pathologists as information specialists. Jama. 2016;316(22):2353–2354.
41. Esteva A, Kuprel B, Novoa RA, Ko J, Swetter SM, Blau HM, et al. Dermatologist-level classification of skin cancer with deep neural networks. Nature. 2017;542(7639):115.
42. Gale W, Oakden-Rayner L, Carneiro G, Bradley AP, Palmer LJ. Detecting hip fractures with radiologist-level performance using deep neural networks. arXiv preprint arXiv:171106504. 2017;.
43. Simmons, M., Singhal, A., & Lu, Z. (2016). Text mining for precision medicine: bringing structure to EHRs and biomedical literature to understand genes and health. In *Translational Biomedical Informatics* (pp. 139-166). Springer, Singapore.
44. Ieraci S, Digiusto E, Sonntag P, Dann L, Fox D. Streaming by case complexity: evaluation of a model for emergency department fast track. Emergency Medicine Australasia. 2008;20(3):241–249.
45. McCoy TH, Yu S, Hart KL, Castro VM, Brown HE, Rosenquist JN, et al. High throughput phenotyping for dimensional psychopathology in electronic health records. Biological psychiatry. 2018;.
46. Forsblom E, Ruotsalainen E, Ollgren J, Järvinen A. Telephone consultation cannot replace bedside infectious disease consultation in the management of Staphylococcus aureus bacteremia. Clinical infectious diseases. 2012;56(4):527–535.
47. Goldstein A, Shahar Y, Orenbuch E, Cohen MJ. Evaluation of an automated knowledge-based textual summarization system for longitudinal clinical data, in the intensive care domain. Artificial intelligence in medicine. 2017;82:20–33.
48. Shneiderman B, Plaisant C, Hesse BW. Improving healthcare with interactive visualization. Computer. 2013;46(5):58–66.
49. Pivovarov R, Coppleson YJ, Gorman SL, Vawdrey DK, Elhadad N. Can Patient Record Summarization Support Quality Metric Abstraction? In: AMIA Annual Symposium Proceedings. vol. 2016. American Medical Informatics Association; 2016. p. 1020.
50. Fishbane S, Niederman MS, Daly C, Magin A, Kawabata M, de Corla-Souza A, et al. The impact of standardized order sets and intensive clinical case management on outcomes in community-acquired pneumonia. Archives of internal medicine. 2007;167(15):1664–1669.
51. Halpern Y. Semi-Supervised Learning for Electronic Phenotyping in Support of Precision Medicine. New York University; 2016.
52. Bates DW, Gawande AA. Improving safety with information technology. New England journal of medicine. 2003;348(25):2526–2534.
53. Shalom E, Shahar Y, Lunenfeld E. An architecture for a continuous, user-driven, and datadriven application of clinical guidelines and its evaluation. Journal of biomedical informatics. 2016;59:130–148.
54. Peleg M, Shahar Y, Quaglini S, Broens T, Budasu R, Fung N, et al. Assessment of a personalized and distributed patient guidance system. International journal of medical informatics. 2017;101:108–130.

55. Kyriacou DN, Anglin D, Taliaferro E, Stone S, Tubb T, Linden JA, et al. Risk factors for injury to women from domestic violence. New England journal of medicine. 1999;341(25):1892–1898.
56. Reis BY, Kohane IS, Mandl KD. Longitudinal histories as predictors of future diagnoses of domestic abuse: modelling study. Bmj. 2009;339:b3677.
57. Landoni G, Comis M, Conte M, Finco G, Mucchetti M, Paternoster G, et al. Mortality in multicenter critical care trials: an analysis of interventions with a significant effect. Critical care medicine. 2015;43(8):1559–1568.
58. McGinnis JM, Stuckhardt L, Saunders R, Smith M. Best care at lower cost: the path to continuously learning health care in America. National Academies Press; 2013.
59. Mills EJ, Thorlund K, Ioannidis JPA. Demystifying trial networks and network meta-analysis. Bmj. 2013;346:f2914.
60. Travers J, Marsh S, Williams M, Weatherall M, Caldwell B, Shirtcliffe P, et al. External validity of randomised controlled trials in asthma: to whom do the results of the trials apply? Thorax. 2007;62(3):219–223.
61. Phillips SP, Hamberg K. Doubly blind: a systematic review of gender in randomised controlled trials. Global health action. 2016;9(1):29597.
62. Hripcsak G, Ryan PB, Duke JD, Shah NH, Park RW, Huser V, et al. Characterizing treatment pathways at scale using the OHDSI network. Proceedings of the National Academy of Sciences. 2016;113(27):7329–7336.
63. Angrist JD, Pischke J. Mostly harmless econometrics: An empiricist's companion. Princeton university press; 2008.
64. Weng SF, Reps J, Kai J, Garibaldi JM, Qureshi N. Can machine-learning improve cardiovascular risk prediction using routine clinical data? PloS one. 2017;12(4):e0174944.
65. Haque A, Guo M, Alahi A, Yeung S, Luo Z, Rege A, et al. Towards Vision-Based Smart Hospitals: A System for Tracking and Monitoring Hand Hygiene Compliance. arXiv preprint arXiv:170800163. 2017;.
66. Council NR. Toward precision medicine: building a knowledge network for biomedical research and a new taxonomy of disease. National Academies Press; 2011.
67. Kellum JA, Prowle JR. Paradigms of acute kidney injury in the intensive care setting. Nature Reviews Nephrology. 2018;.
68. Alge JL, Arthur JM. Biomarkers of AKI: a review of mechanistic relevance and potential therapeutic implications. Clinical Journal of the American Society of Nephrology. 2014;.
69. Schulam P, Saria S. A framework for individualizing predictions of disease trajectories by exploiting multi-resolution structure. In: Neural Information Processing Systems (NIPS); 2015. .
70. Xu Y, Xu Y, Saria S. A Bayesian nonparametric approach for estimating individualized treatment-response curves. In: Machine Learning for Healthcare Conference (MLHC); 2016. .
71. Araujo A, Julious S, Senn S. Understanding variation in sets of N-of-1 trials. PloS one. 2016;.
72. Park ST, Chu W. Pairwise preference regression for cold-start recommendation. In: ACM Conference on Recommender Systems. ACM; 2009.
73. Chen, R., & Snyder, M. (2012). Systems biology: personalized medicine for the future?. *Current opinion in pharmacology*, *12*(5), 623-628.
74. Tian, Q., Price, N. D., & Hood, L. (2012). Systems cancer medicine: towards realization of predictive, preventive, personalized and participatory (P4) medicine. *Journal of internal medicine*, *271*(2), 111-121.
75. Chen, Rui, and Michael Snyder. "Promise of personalized omics to precision medicine." *Wiley Interdisciplinary Reviews: Systems Biology and Medicine* 5.1 (2013): 73-82.
76. Van Bebber, S. L., Trosman, J. R., Liang, S. Y., Wang, G., Marshall, D. A., Knight, S., & Phillips, K. A. (2010). Capacity building for assessing new technologies: approaches to examining personalized medicine in practice. *Personalized medicine*, *7*(4), 427-439.
77. Gabriel Sanchez R, Sanchez Gomez LM, Carmona L, Roqué i Figuls M, Bonfill Cosp X. Hormone replacement therapy for preventing cardiovascular disease in post-menopausal women. The Cochrane Library. 2005;.
78. Prentice RL, Langer RD, Stefanick ML, Howard BV, Pettinger M, Anderson GL, et al. Combined analysis of Women's Health Initiative observational and clinical trial data on postmenopausal hormone treatment and cardiovascular disease. American Journal of Epidemiology. 2006;163(7):589–599.
79. Tsymbal A. The problem of concept drift: definitions and related work. Computer Science Department, Trinity College Dublin. 2004;106(2).
80. Lazer D, Kennedy R, King G, Vespignani A. The parable of Google Flu: traps in big data analysis. Science. 2014;343(6176):1203–1205.
81. Gong JJ, Naumann T, Szolovits P, Guttag JV. Predicting Clinical Outcomes Across Changing Electronic Health Record Systems. In: International Conference on Knowledge Discovery and Data Mining (KDD). ACM; 2017. p. 1497–1505.
82. Nestor, Bret, et al. "Feature robustness in non-stationary health records: caveats to deployable model performance in common clinical machine learning tasks." *arXiv preprint arXiv:1908.00690* (2019).
83. Subbaswamy A, Saria S. Counterfactual normalization: proactively addressing dataset shift using causal mechanisms. In: Uncertainty in Artificial Intelligence (UAI); 2018. p. 947–957.

84. Subbaswamy A, Schulam P, Saria S. Preventing Failures Due to Dataset Shift: Learning Predictive Models That Transport. In: Artificial Intelligence and Statistics (AISTATS); 2019. .
85. Lum K, Isaac W. To predict and serve? Significance. 2016;13(5):14–19.
86. Dressel J, Farid H. The accuracy, fairness, and limits of predicting recidivism. Science Advances. 2018;4(1):eaao5580.
87. Hysong SJ, Best RG, Pugh JA. Audit and feedback and clinical practice guideline adherence: making feedback actionable. Implementation Science. 2006;1(1):9.
88. Obermeyer Z, Lee TH. lost in Thought—The Limits of the Human Mind and the Future of Medicine. New England Journal of Medicine. 2017;377(13):1209–1211.
89. Manrai AK, Bhatia G, Strymish J, Kohane IS, Jain SH. Medicine's uncomfortable relationship with math: calculating positive predictive value. JAMA internal medicine. 2014;174(6):991– 993.
90. Doshi-Velez F, Kim B. Towards a rigorous science of interpretable machine learning. 2017;.
91. Letham B, Rudin C, McCormick TH, Madigan D, et al. Interpretable classifiers using rules and bayesian analysis: Building a better stroke prediction model. The Annals of Applied Statistics. 2015;9(3):1350–1371.
92. Ribeiro MT, Singh S, Guestrin C. Why should I trust you?: Explaining the predictions of any classifier. In: International Conference on Knowledge Discovery and Data Mining (KDD). ACM; 2016. p. 1135–1144.
93. Koh PW, Liang P. Understanding black-box predictions via influence functions; 2017. .
94. Finlayson SG, Kohane IS, Beam AL. Adversarial Attacks Against Medical Deep Learning Systems. arXiv preprint arXiv:180405296. 2018;.
95. Han X, Hu Y, Foschini L, Chintz L, Jankelson L, Ranganath R. Adversarial Examples for Electrocardiograms. arXiv preprint arXiv:190505163. 2019;.
96. Charon R. Narrative medicine: a model for empathy, reflection, profession, and trust. Journal of the American Medical Association. 2001;286(15):1897–1902.
97. Stewart M. Patient-centered medicine: transforming the clinical method. Radcliffe Publishing; 2003.
98. Settles B. Active learning. Synthesis Lectures on Artificial Intelligence and Machine Learning. 2012;6(1):1–114.
99. Abbeel P, Ng AY. Apprenticeship learning via inverse reinforcement learning. In: Proceedings of the twenty-first international conference on Machine learning. ACM; 2004. p. 1.
100. Cho S. Towards creative evolutionary systems with interactive genetic algorithm. Applied Intelligence. 2002;16(2):129–138.
101. Zhu J, Krähenbühl P, Shechtman E, Efros AA. Generative visual manipulation on the natural image manifold. In: European Conference on Computer Vision. Springer; 2016. p. 597–613.
102. Beam AL, Kohane IS. Big data and machine learning in health care. JAMA. 2018;Available from: +http://dx.doi.org/10.1001/jama.2017.18391.
103. Chen ML, Doddi A, Royer J, Freschi L, Schito M, Ezewudo M, et al. Deep Learning Predicts Tuberculosis Drug Resistance Status from Whole-Genome Sequencing Data. bioRxiv. 2018;p. 275628.
104. Ranganath R, Perotte A, Elhadad N, Blei D. Deep Survival Analysis. In: Machine Learning for Healthcare Conference; 2016. p. 101–114.
105. Harutyunyan H, Khachatrian H, Kale DC, Galstyan A. Multitask Learning and Benchmarking with Clinical Time Series Data. arXiv preprint arXiv:170307771. 2017;.
106. Poplin R, Varadarajan AV, Blumer K, Liu Y, McConnell MV, Corrado GS, et al. Prediction of cardiovascular risk factors from retinal fundus photographs via deep learning. Nature Biomedical Engineering. 2018;p. 1.
107. Che Z, Purushotham S, Cho K, Sontag D, Liu Y. Recurrent neural networks for multivariate time series with missing values. Scientific reports. 2018;8(1):6085.
108. Bengio Y, Courville A, Vincent P. Representation learning: A review and new perspectives. IEEE transactions on pattern analysis and machine intelligence. 2013;35(8):1798–1828.
109. Socher R, Ganjoo M, Manning CD, Ng A. Zero-shot learning through cross-modal transfer. In: Advances in Neural Information Processing Systems (NIPS); 2013. p. 935–943.
110. Nickel M, Kiela D. Poincare Embeddings for Learning Hierarchical Representations. arXiv preprint arXiv:170508039. 2017;.
111. Choi E, Bahadori MT, Song L, Stewart WF, Sun J. GRAM: Graph-based attention model for healthcare representation learning. In: International Conference on Knowledge Discovery and Data Mining (KDD). ACM; 2017. p. 787–795.
112. Lin C, Jain S, Kim H, Bar-Joseph Z. Using neural networks for reducing the dimensions of single-cell RNA-Seq data. Nucleic acids research. 2017;45(17):e156–e156.
113. Shah, Nigam H., Arnold Milstein, and Steven C. Bagley. "Making machine learning models clinically useful." *Jama* 322.14 (2019): 1351-1352